\pdfoutput=1
\documentclass[11pt]{article}

\usepackage{acl}

\usepackage{times}
\usepackage{latexsym}

\usepackage[T1]{fontenc}
\usepackage[utf8]{inputenc}

\usepackage{microtype}
\usepackage{inconsolata}
\usepackage{kotex}
\usepackage{graphicx}
\usepackage{amsmath,amssymb}
\usepackage{amsfonts,amsthm,mathtools}
\usepackage{algorithm}
\usepackage{algorithmicx}
\usepackage{caption}
\usepackage{footnote}
\usepackage{tabularx}
\usepackage{dsfont}
\usepackage{multirow}

\usepackage{listings}[breaklines=true]
\lstdefinestyle{mystyle}{
    backgroundcolor=\color{gray!10},   
    commentstyle=\color{green},
    keywordstyle=\color{blue},
    numberstyle=\tiny\color{gray},
    stringstyle=\color{purple},
    basicstyle=\ttfamily\footnotesize,
    breakatwhitespace=false,         
    breaklines=true,                 
    captionpos=b,                    
    keepspaces=true,                 
    numbers=left,                    
    numbersep=5pt,                  
    showspaces=false,                
    showstringspaces=false,
    showtabs=false,                  
    tabsize=2
}
\usepackage{booktabs} % For better table lines
\usepackage{multirow} % For multirow feature
\usepackage{multicol} % For multicolumn feature

\usepackage{comment} 
\usepackage{subcaption}
\title{Efficient and Effective Vocabulary Expansion \\Towards Multilingual Large Language Models}

\author{
    \quad Seungduk Kim\thanks{\quad Equal Contribution.}
    \quad Seungtaek Choi\footnotemark[1]
    \quad \textbf{Myeongho Jeong} \\
    {Yanolja, South Korea} \\
    {\tt \{seungduk.kim, seungtaek.choi, myeongho.jeong\}@yanolja.com} \\
}

\begin{document}
\maketitle
\newtheorem{example}{Example}

\newcommand\Tstrut{\rule{0pt}{2.2ex}}       % "top" strut
\newcommand\Bstrut{\rule[-0.6ex]{0pt}{0pt}} % "bottom" strut
\newcommand{\TBstrut}{\Tstrut\Bstrut} % top&bottom struts

\newcommand{\todoc}[2]{{\textcolor{#1}{\textbf{#2}}}}
\newcommand{\todored}[1]{\todoc{red}{\textbf{#1}}}
\newcommand{\todoblue}[1]{\todoc{blue}{\textbf{[[#1]]}}}
\newcommand{\todogreen}[1]{\todoc{green}{\textbf{[[#1]]}}}
\newcommand{\todoorange}[1]{\todoc{orange}{\textbf{[[#1]]}}}
\newcommand{\todopurple}[1]{\todoc{purple}{\textbf{[[#1]]}}}
\newcommand{\todobrown}[1]{\todoc{brown}{\textbf{[[#1]]}}}

\newcommand{\kjh}[1]{\todored{jihyuk: #1}}
\newcommand{\mj}[1]{\todoblue{MJ: #1}}
\newcommand{\yj}[1]{\todoblue{YJ: #1}}
\newcommand{\sw}[1]{\todoorange{SW: #1}}
\newcommand{\hist}[1]{\todoorange{hist: #1}}
\newcommand{\mh}[1]{\todobrown{mh: #1}}
\newcommand{\ms}[1]{\todopurple{MS: #1}}

\newcommand{\se}{{\it SE}}%
\newcommand{\eg}{{\it e.g.}}%
\newcommand{\ie}{{\it i.e.}}%
\newcommand{\etal}{{\it et al.}}%
\newcommand{\etc}{{\it etc}}%
\newcommand{\ours}{{WISE}}%

\newcommand{\argmin}{\operatornamewithlimits{argmin}}
\newcommand{\argmax}{\operatornamewithlimits{argmax}}
\definecolor{yellow-green}{rgb}{0.3, 0.5, 0.0}

\def\geotextual{{spatial-keyword}}
\def\geospatial{geo-spatial}
\def\PI{\mathcal{P}}
\newcommand{\XXP}[1]{{\PI(#1)}}
\def\XXQEO{\emph{$Q_1$}}
\def\kNN{\textsc{$k$NN}}
\def\XXD{\mathcal{D}}
\def\XXT{\mathcal{T}}
\newcommand{\XXDN}[0]{{D}}
\newcommand{\XXTN}[0]{{T}}
\def\Base{\textsc{Base}}
\def\TopK{\textsc{Top-$k$}}
\def\tag{{keyword}}
\def\Query{{Query}}
\newcommand{\ttag}[1]{{`#1'}}

\newcommand{\base}{\textsf{NER}}
\newcommand{\baseHash}{\textsf{NER+Hash}}

% Caligraphy style
\newcommand{\mcal}[1]{{\cal{#1}}}
\newcommand{\calA}{\mbox{${\cal A}$}}
\newcommand{\calB}{\mbox{${\cal B}$}}
\newcommand{\calC}{\mbox{${\cal C}$}}
\newcommand{\calD}{\mbox{${\cal D}$}}
\newcommand{\calE}{\mbox{${\cal E}$}}
\newcommand{\calF}{\mbox{${\cal F}$}}
\newcommand{\calG}{\mbox{${\cal G}$}}
\newcommand{\calH}{\mbox{${\cal H}$}}
\newcommand{\calI}{\mbox{${\cal I}$}}
\newcommand{\calJ}{\mbox{${\cal J}$}}
\newcommand{\calK}{\mbox{${\cal K}$}}
\newcommand{\calL}{\mbox{${\cal L}$}}
\newcommand{\calM}{\mbox{${\cal M}$}}
\newcommand{\calN}{\mbox{${\cal N}$}}
\newcommand{\calO}{\mbox{${\cal O}$}}
\newcommand{\calP}{\mbox{${\cal P}$}}
\newcommand{\calQ}{\mbox{${\cal Q}$}}
\newcommand{\calR}{\mbox{${\cal R}$}}
\newcommand{\calS}{\mbox{${\cal S}$}}
\newcommand{\calT}{\mbox{${\cal T}$}}
\newcommand{\calU}{\mbox{${\cal U}$}}
\newcommand{\calV}{\mbox{${\cal V}$}}
\newcommand{\calW}{\mbox{${\cal W}$}}
\newcommand{\calX}{\mbox{${\cal X}$}}
\newcommand{\calY}{\mbox{${\cal Y}$}}
\newcommand{\calZ}{\mbox{${\cal Z}$}}
\begin{abstract}

This report introduces \texttt{EEVE-Korean-v1.0}, a Korean adaptation of large language models that exhibit remarkable capabilities across English and Korean text understanding. Building on recent highly capable but English-centric LLMs, such as SOLAR-10.7B and Phi-2, where non-English texts are inefficiently processed with English-centric tokenizers, we present an efficient and effective vocabulary expansion (EEVE) method, which encompasses parameter freezing and subword initialization. 
In contrast to previous efforts that believe new embeddings require trillions of training tokens, we show that our method can significantly boost non-English proficiency within just 2 billion tokens. 
Surpassing most instruction-tuned LLMs on the Open Ko-LLM Leaderboard, as of January 2024, our model \texttt{EEVE-Korean-10.8B-v1.0} ranks as the leading Korean pre-trained model in the open-source community, according to Hugging Face's leaderboard. 
We open-source our models on Huggingface to empower the open research community in various languages. 

\end{abstract}
\section{Introduction}
\label{sec:introduction}

Recent advancements in the field of large language models (LLMs), such as GPT-4~\cite{openai-gpt4}, Gemini~\cite{google-gemini}, and Claude~\cite{anthropic-claude}, have demonstrated remarkable capabilities in processing and understanding multiple languages. 
On the other hand, though notable models in open source community, such as LLaMA~\cite{touvron2023llama,touvron2023llama2}, MPT~\cite{mosaicml2023mpt}, Falcon~\cite{almazrouei2023falcon}, Mistral~\cite{jiang2023mistral}, Mixtral~\cite{jiang2024mixtral}, SOLAR~\cite{kim2023solar}, and Phi-1.5~\cite{li2023textbooks} have set benchmarks in English tasks, these developments have predominantly favored English, leading to a performance gap in non-English languages. 

Such disparity can be found not only in their language proficiency but also in computational efficiency, where non-English languages like Korean require significantly more tokens than English even for equivalent semantic content (Figure~\ref{tab:tokenizer}). 
And, of course, this negatively affects the user experiences, such as longer response times, shorter context lengths, and higher API costs~\cite{petrov2023language}. 
Expanding the tokenizer vocabulary, which introduces some frequently used yet long words as additional tokens, is thus indispensable for non-English users, but vocabulary expansion is a very challenging task because new embeddings require trillions of training tokens~\cite{zhao2024llama}. 

To this end, this technical report presents a novel approach for \textbf{e}fficient and \textbf{e}ffective \textbf{v}ocabulary \textbf{e}xpansion, namely EEVE, which can better train the embeddings of newly added tokens. 
For ease of adaptation, we utilize subword-based embedding initialization and design seven training stages with parameter freezing, which elaborately adjust the order and amount of parameters to be trained. 
We meticulously transfer the advanced capabilities of foundational models from English to Korean by initially focusing on the training of only input embeddings and progressively expanding to encompass the full parameters in the final stage.

Using \texttt{EEVE}, we officially release a family of Korean LLMs, \texttt{EEVE-Korean-10.8B-v1.0}\footnote{\url{https://huggingface.co/yanolja/EEVE-Korean-10.8B-v1.0}} and \texttt{EEVE-Korean-2.8B-v1.0}\footnote{\url{https://huggingface.co/yanolja/EEVE-Korean-2.8B-v1.0}}, which are built on recent English-centric LLMs, specifically SOLAR-10.7B~\cite{kim2023solar} and Phi-2~\cite{li2023textbooks}, with further Korean-centric pre-training. 
We evaluate our models on \texttt{lm-evaluation-harness}\footnote{\url{https://github.com/EleutherAI/lm-evaluation-harness}}~\cite{eval-harness} for both English and Korean language tasks, such as boolean question answering (BoolQ;~\citealt{clark2019boolq}), commonsense causal reasoning (COPA;~\citealt{roemmele2011choice}, context-sensitive word understanding (WiC;~\citealt{pilehvar2019wic}), commonsense reasoning (HellaSwag;~\citealt{zellers2019hellaswag}), and sentiment negation recognition (SentiNeg). 
From the evaluation, we observe that our models outperform the recent open Korean pre-trained LLMs like OPEN-SOLAR-KO-10.7B~\cite{solar_ko_junbum_2023}, Polyglot-Ko~\cite{ko2023technical}, and KoGPT~\cite{KoGPT2021}, while preserving the strong English capability of the base English-centric LLMs in terms of benchmark performance, being ranked as the leading Korean pre-trained model in Open Ko-LLM Leaderboard~\cite{open-ko-llm-leaderboard}.
\section{Efficient and Effective Vocabulary Expansion}
\label{sec:approach}

\begin{figure*}
    \begin{subfigure}[b]{0.24\textwidth} % 각 그림의 너비를 전체 텍스트 너비의 14%로 설정
        \centering
        \includegraphics[height=3.6cm]{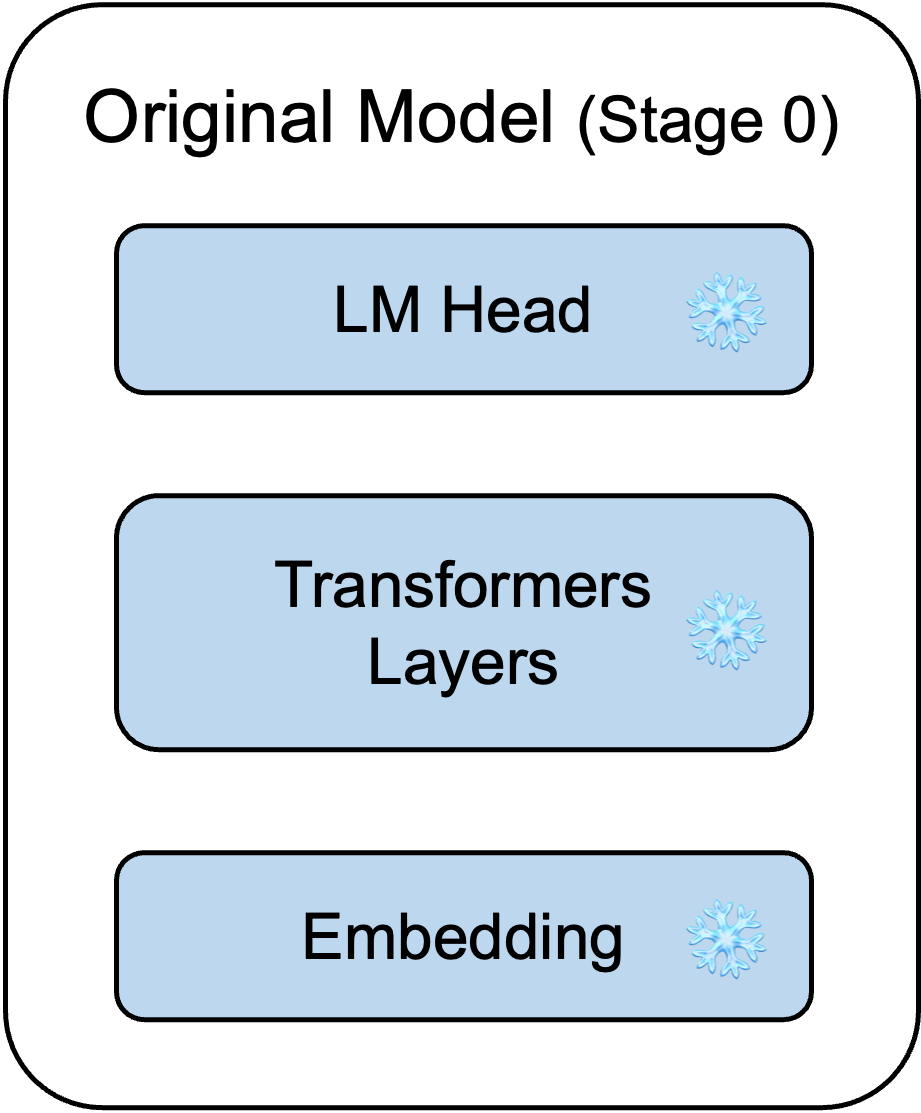}
        \caption{Stage 0}
    \end{subfigure}
    \hfill % 가로로 나란히 배열
    \begin{subfigure}[b]{0.24\textwidth}
        \centering
        \includegraphics[height=3.6cm]{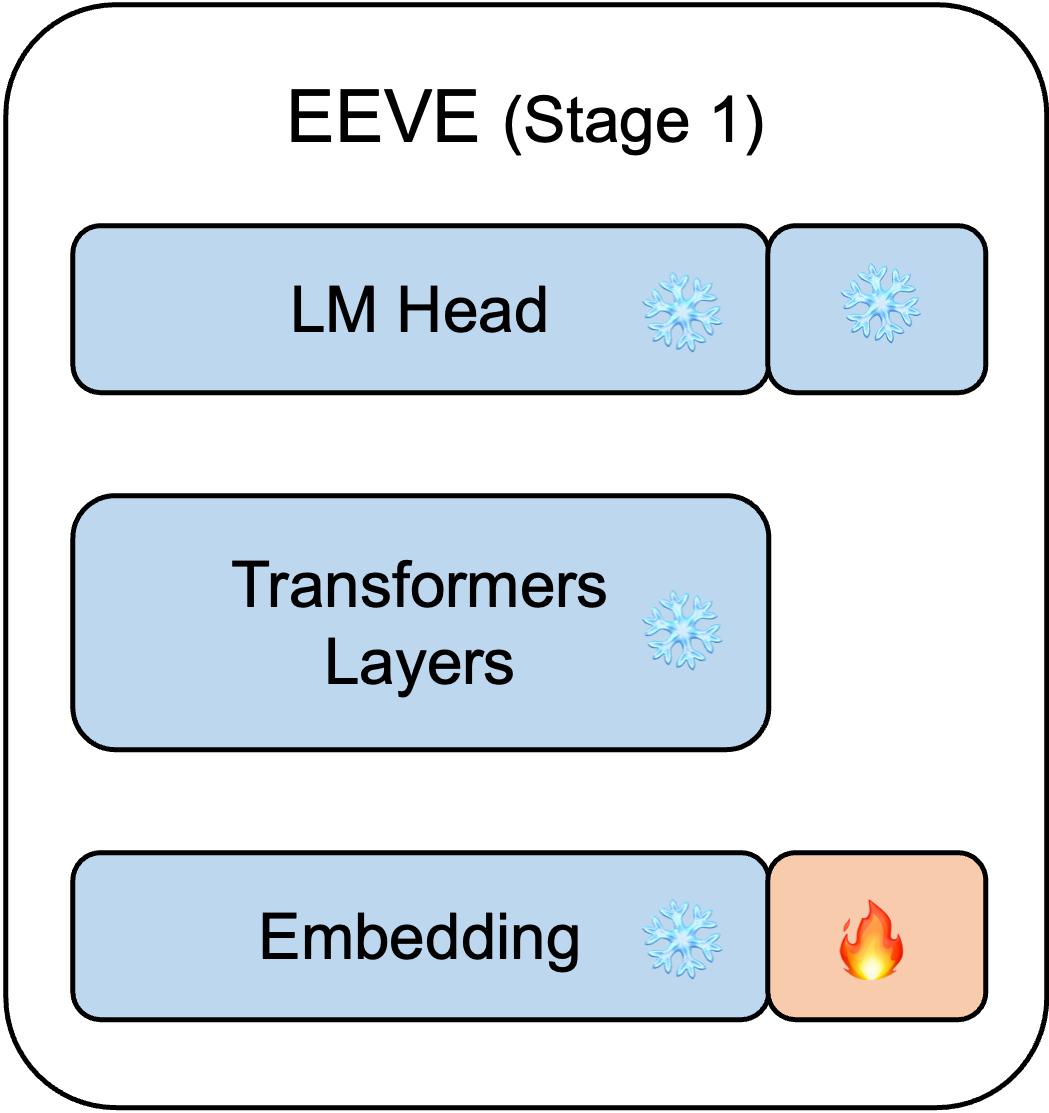}
        \caption{Stage 1}
    \end{subfigure}
    \hfill
    \begin{subfigure}[b]{0.24\textwidth}
        \centering
        \includegraphics[height=3.6cm]{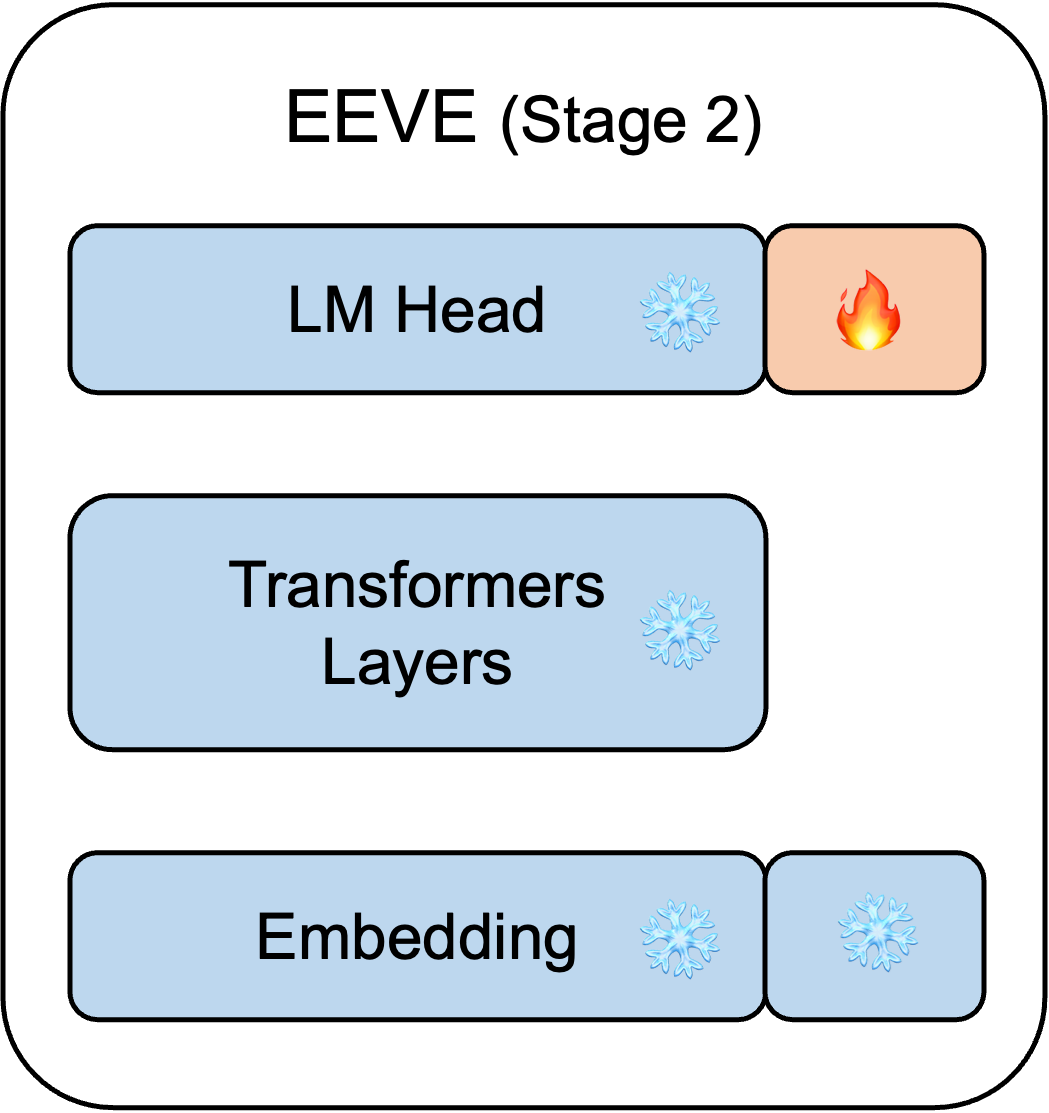}
        \caption{Stage 2}
    \end{subfigure}
    \hfill
    \begin{subfigure}[b]{0.24\textwidth}
        \centering
        \includegraphics[height=3.6cm]{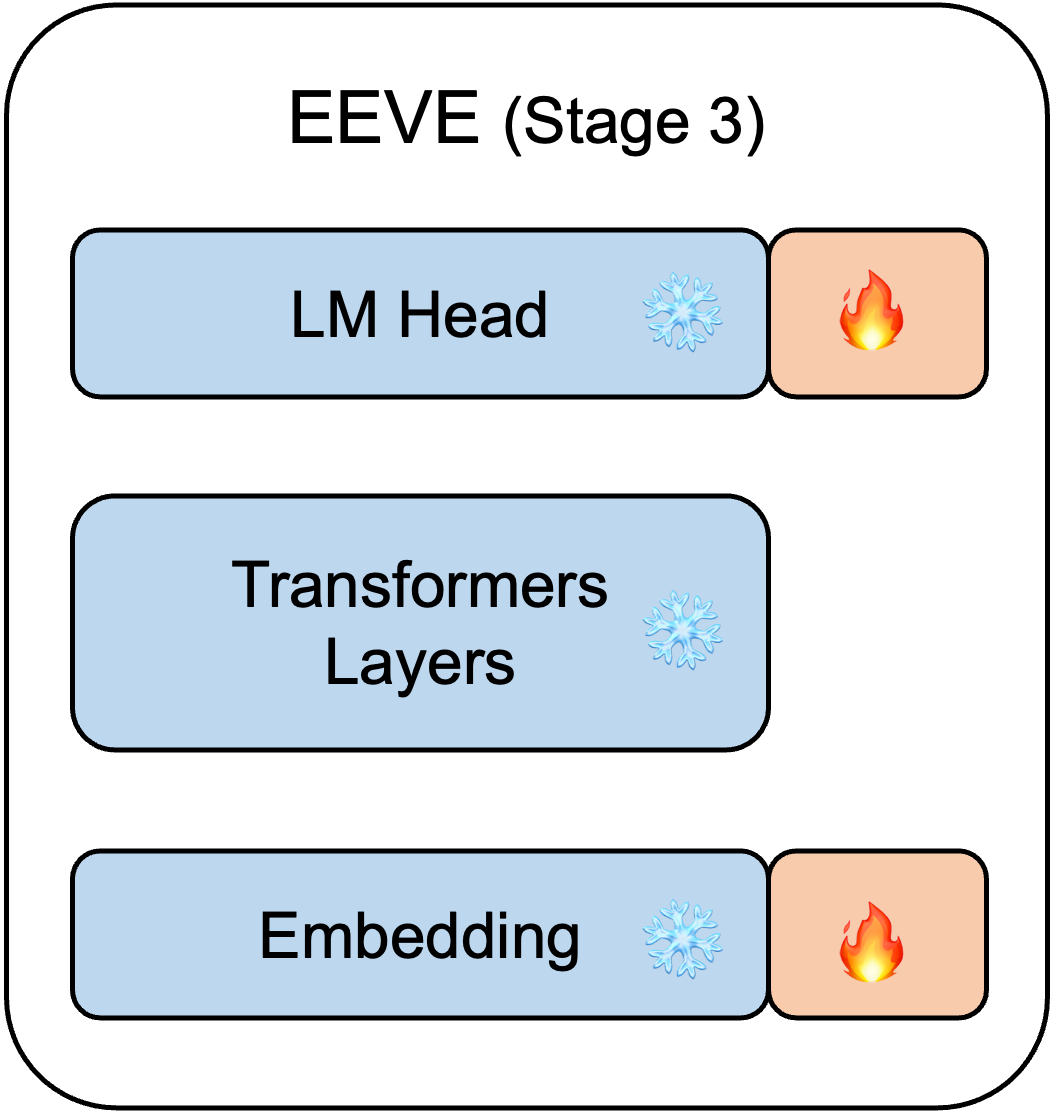}
        \caption{Stage 3}
    \end{subfigure}
    \\
    \begin{subfigure}[b]{0.24\textwidth}
        \centering
        \includegraphics[height=3.6cm]{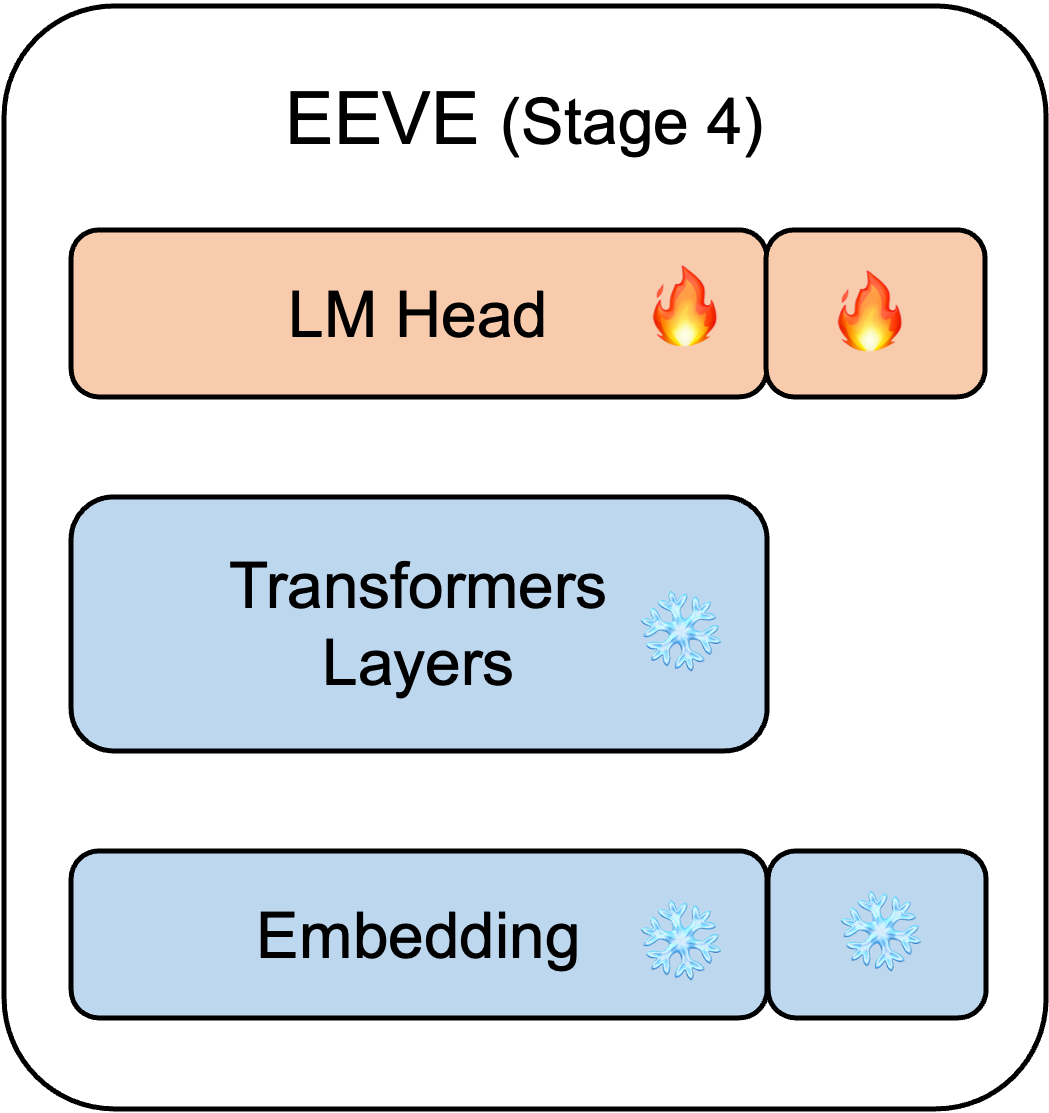}
        \caption{Stage 4}
    \end{subfigure}
    \hfill
    \begin{subfigure}[b]{0.24\textwidth}
        \centering
        \includegraphics[height=3.6cm]{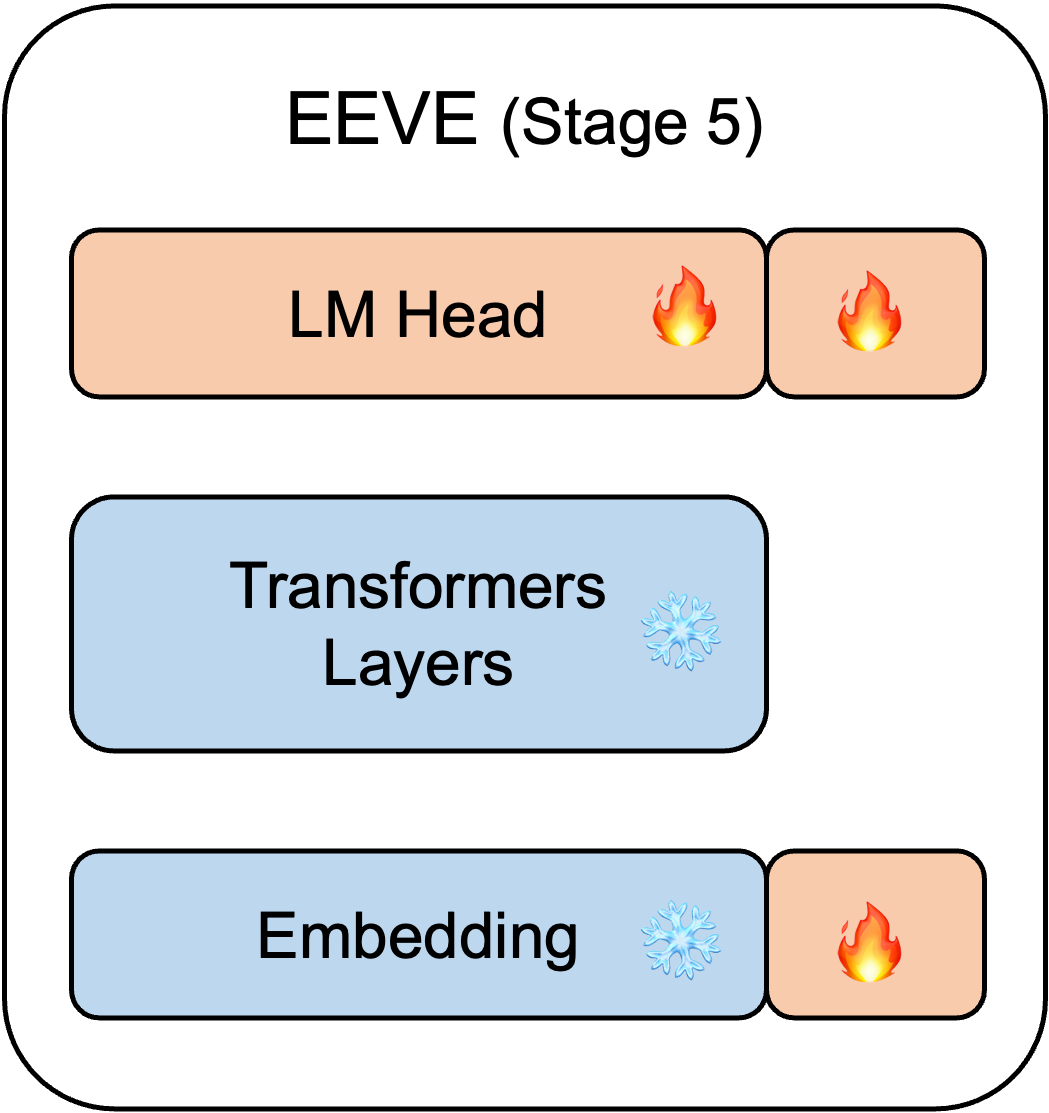}
        \caption{Stage 5}
    \end{subfigure}
    \hfill
    \begin{subfigure}[b]{0.24\textwidth}
        \centering
        \includegraphics[height=3.6cm]{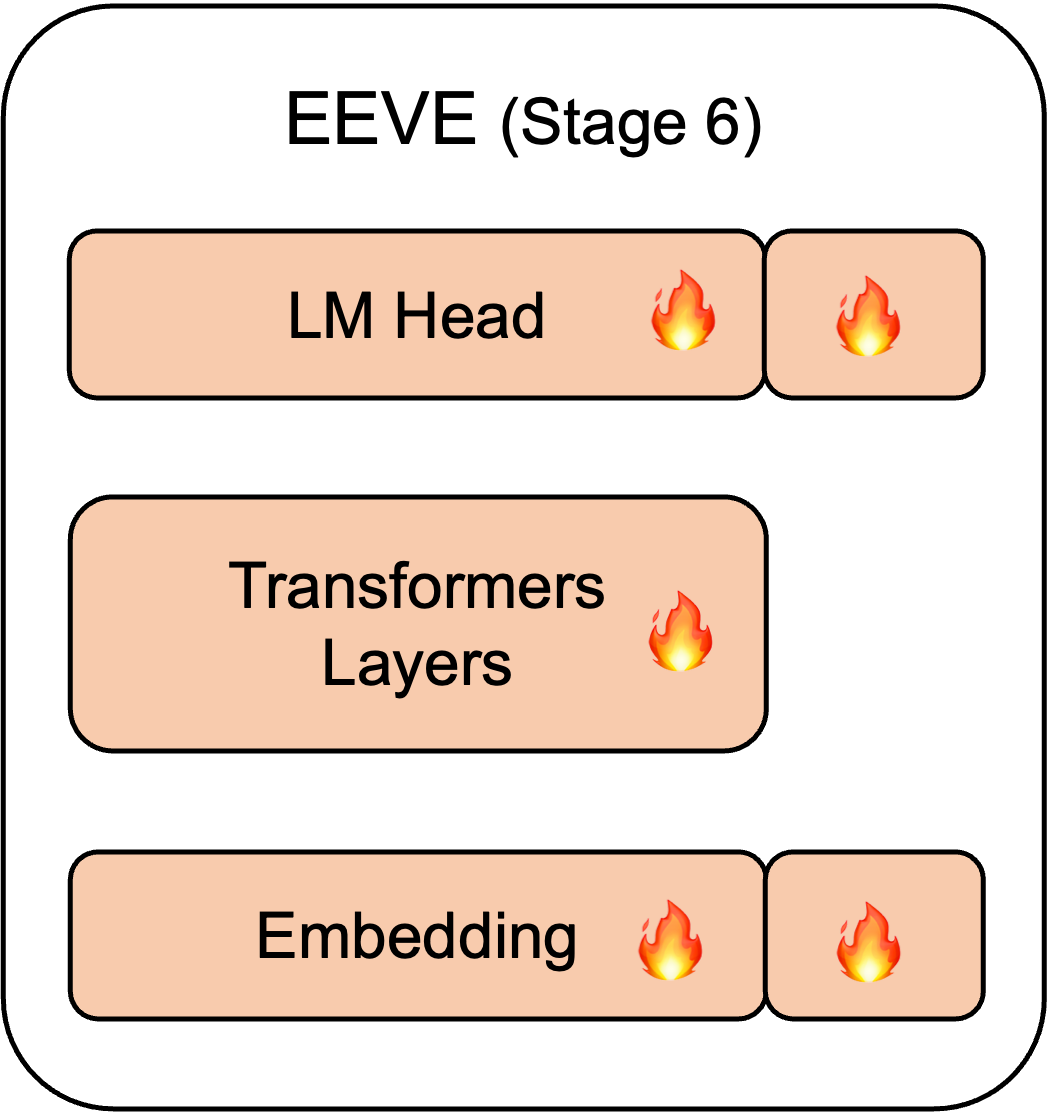}
        \caption{Stage 6}
    \end{subfigure}
    \begin{subfigure}[b]{0.24\textwidth}
        \centering
        \includegraphics[height=3.6cm]{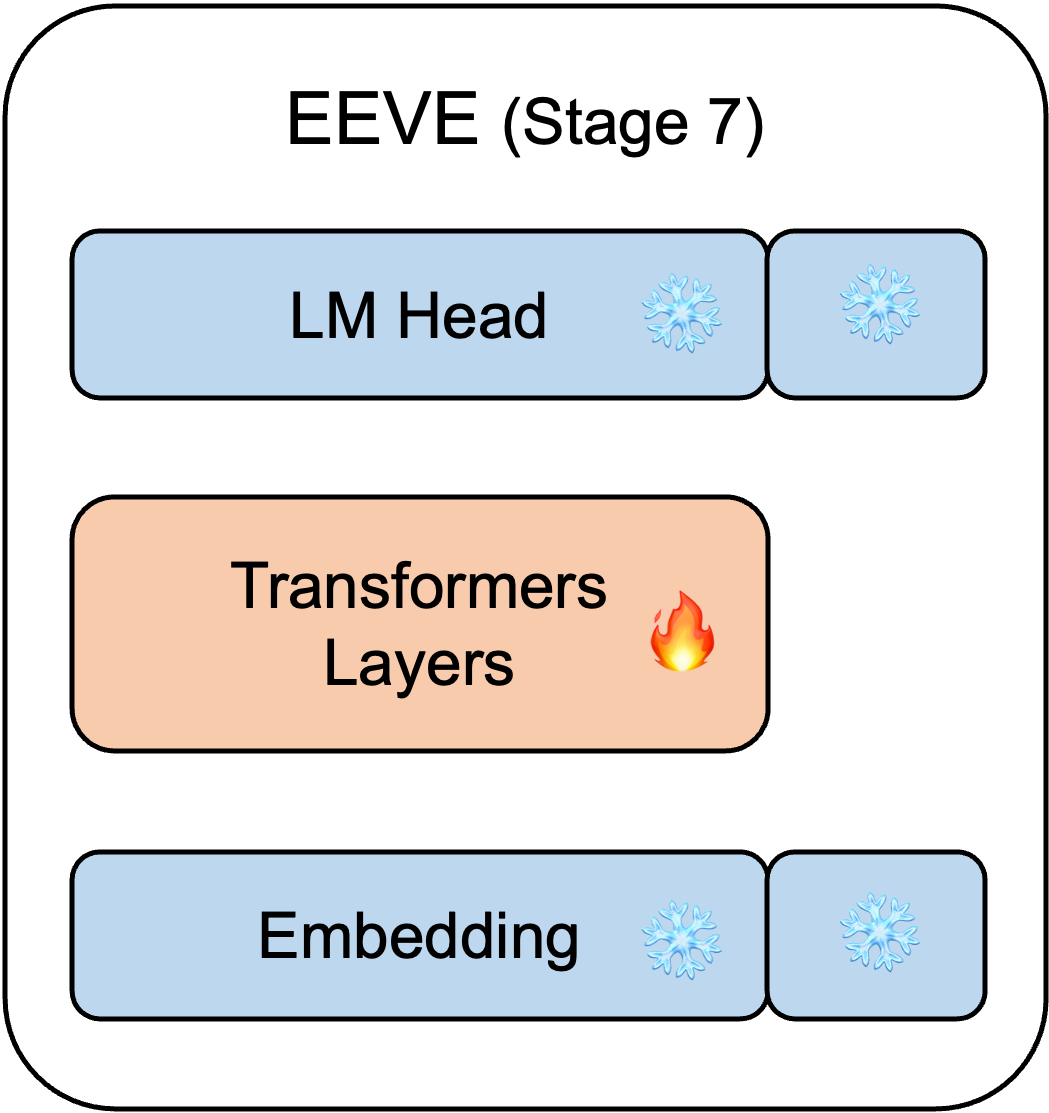}
        \caption{Stage 7}
    \end{subfigure}
    \caption{Training stages with parameter freezing. The fire and snowflake emojis indicate the trainable and frozen parameters respectively.}
    \label{fig:stages}
\end{figure*}

To address the challenge of efficiently extending English-centric Language Models (LLMs) to include non-English languages, we introduce a novel methodology for vocabulary expansion. This method combines parameter freezing with subword-based embedding initialization to effectively incorporate and adapt to new linguistic tokens from languages beyond its initial training scope, thereby enhancing its applicability across various linguistic contexts. Our approach outlines a structured seven-stage training process, as illustrated in Figure~\ref{fig:stages}, meticulously designed to effectively integrate new tokens into the model's vocabulary. During pre-training, our objective is causal language modeling.

Our core assumption is that foundational models, having been extensively trained in English texts, possess a substantial level of understanding and reasoning capabilities. Transferring these capabilities from English to another language, such as Korean, could be more efficient than developing performance from standalone Korean pre-training.

\subsection{Preliminary 1: Tokenizer Training}
\begin{table}[t!]
  \centering
  \resizebox{0.92\columnwidth}{!}{
  \begin{tabular}{p{0.95\columnwidth}}
    \hline \noalign{\hrule height0.8pt} %%%%=
    \textbf{English (\textit{8 tokens})} \\
    ``\textit{Hello, the weather is nice today.}'' \\
    {\small \texttt{[`\textunderscore Hello', `,', `\textunderscore the', `\textunderscore weather', `\textunderscore is', `\textunderscore nice', `\textunderscore today', `.'] }} \\
    
    \hline \noalign{\hrule height0.8pt} %%%%=
    \textbf{Korean (\textit{26 tokens})} \\
    ``\textit{안녕하세요, 오늘은 날씨가 좋네요.}'' \\
    {\small \texttt{[`\textunderscore', `안', `$<$0xEB$>$', `$<$0x85$>$', `$<$0x95$>$', `하', `세', `요', `,', `\textunderscore', `오', `$<$0xEB$>$', `$<$0x8A$>$', `$<$0x98$>$', `은', `\textunderscore', `날', `$<$0xEC$>$', `$<$0x94$>$', `$<$0xA8$>$', `가', `\textunderscore', `좋', `네', `요', `.'] }} \\
    
    \hline \noalign{\hrule height0.8pt} %%%%=
    \textbf{Expanded Tokenizer (\textit{9 tokens})} \\
    {\small \texttt{[`\textunderscore안', `녕', `하세요', `,', `\textunderscore오늘은', `\textunderscore날씨가', `\textunderscore좋', `네요', `.'] }} \\
    \hline \noalign{\hrule height0.8pt} %%%%=
  \end{tabular}
  }
  \caption{A comparison of token consumption between English and Korean. We used the tokenizers of SOLAR~\cite{kim2023solar} and our \texttt{EEVE-Korean-10.8B-v1.0}.}
  \label{tab:tokenizer}
\end{table}

We trained a new tokenizer on our Korean corpus. Since our goal is to maximize the leverage of the base model's performance, we maintained the base model's vocabulary and added 8,960 tokens from the corpus that appeared at least 6,000 times, prioritizing those with the highest frequency. Ultimately, the tokenizer's vocabulary expanded to 40,960 tokens for \texttt{EEVE-Korean-10.8B-v1.0}. This process entailed several rounds of tokenizer training and a manual curation of tokens, based on an analysis of token frequency, ensuring a comprehensive and relevant vocabulary for our model. 
As shown in Table~\ref{tab:tokenizer}, the overall token consumption for Korean texts is significantly improved, almost three-fold, contributing to the reduction of computational costs during the entire training process. 
% More details can be found in Appendix~\ref{appendix:tokenizer}. 

\subsection{Preliminary 2: Subword-based Embeddings Initialization} 
\label{sec:initialization}
The integration process starts before actual training, introducing new input and output embeddings, called \texttt{embed\_tokens} and \texttt{lm\_head}, to the model's parameters. This preliminary step is crucial for preparing for the sophisticated learning process that follows.

For the input embeddings of the newly added tokens, we adopt the approach of using the average embeddings of the subword tokens that make up these new tokens as in~\cite{hewitt2021initializing,welch2020improving}. This method utilizes the semantic richness of the model's existing subword embeddings to offer a meaningful starting point for the new tokens' representations.

Conversely, the output embeddings for the newly added tokens are initialized with the embeddings of the first subword token that comprises the new token. This strategy aims to align the new tokens' output representations closely with the semantic characteristics of their constituent subwords, enabling a smoother integration into the model's predictive framework. The significance of such initialization will be further discussed.

\subsection{Multi-stage Training}

Here we describe the nuanced approach of our seven-stage training methodology for efficient vocabulary expansion, emphasizing the meticulous process of integrating new tokens derived from languages beyond the initial English-centric training scope.

\begin{figure}[t]
    \centering
    \includegraphics[width=0.75\columnwidth]{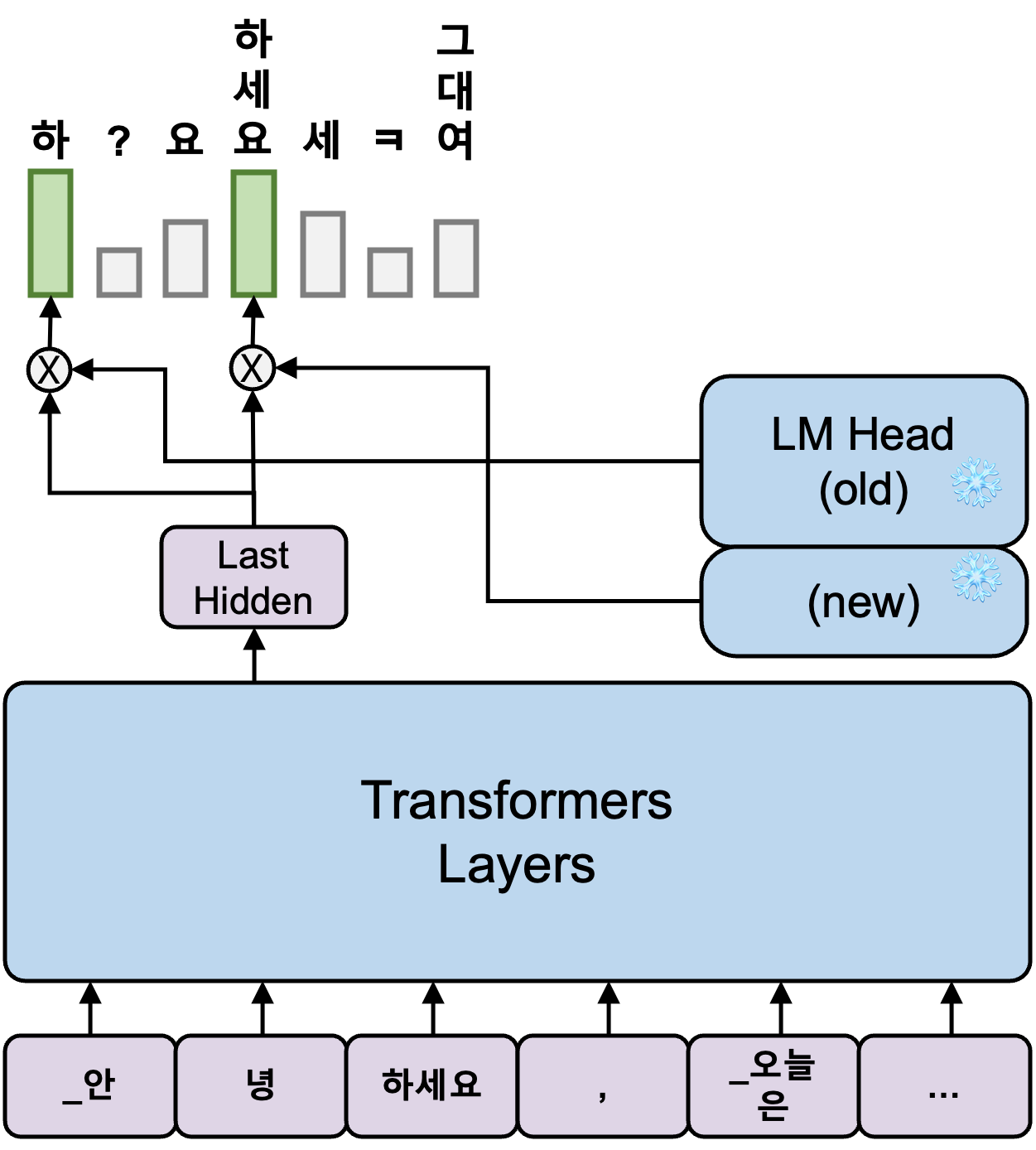}
    \caption{An illustrative example of showing how our subword-based embedding initialization enables harmonize the old and new tokens at Stage 1. In Stage 1, the output embeddings of newly added tokens are initialized with the output embeddings of their first subword tokens that make up these new tokens, such that the last hidden representation for predicting ``하세요'' yields the same logits for the newly added token ``하세요'' with its first subword token ``하''. Even if we give the new token ``하세요'' as a gold token, the gradients are eventually computed based on its subword token ``하'', so the model takes the input embeddings of ``하세요'' to predict its subword ``하''.}
    \label{fig:subword}
\end{figure}

\noindent\textbf{Stage 1 (new input embeddings):} Initially, our focus is narrow yet critical: to learn the input embeddings of the newly added tokens while freezing all other model parameters. This stage is foundational, allowing the model to adjust its recognition and processing of these tokens from the beginning. The pre-initialized embeddings serve as a starting point, guiding the model to better utilize these new tokens in its existing framework. Our principal hypothesis here is that if the input and output token sequences in causal language modeling can be differentiated, by utilizing both the old and new tokenizers at the same time, the model can more efficiently and effectively learn new vocabulary embeddings, as it could leverage its established knowledge in the embedding spaces from old tokens. However, employing distinct tokenizers for input and output sequences at once poses implementation challenges, such as the difficulty of applying teacher forcing due to mismatched input/output sequences. Here, the subword-based embedding initialization (Sec~\ref{sec:initialization}) provides a proxy for using the old tokenizer for output sequences, such that the model is tasked to generate the subword token (old) given the whole word token (new).
In other words, the model could learn to align their representations for generating the new token with that for generating its first subword token, by optimizing only the input embeddings without any modification of input/output token sequences as described in Figure~\ref{fig:subword}. 
However, at this stage, the model is not yet able to distinguish between tokens sharing the same hidden state.

\noindent\textbf{Stage 2 (new output embeddings):} 
Our goal is to enhance the model's proficiency in accurately generating new tokens across various contexts by solely adjusting the output embeddings (\texttt{lm\_head}). The decision to freeze all other parameters stems from the model's current unstable state. Allowing both input and output embeddings to be trained simultaneously would complicate achieving convergence, thus hindering the model's progress toward optimal performance. By freezing most of the parameters, we achieve more stable convergence. Moreover, this approach significantly reduces the training time, as it eliminates the necessity for backpropagation through the other layers.

\noindent\textbf{Stage 3 (new input and output embeddings):} 
At this stage, the input embeddings (\texttt{embed\_tokens}) still remain optimized based on the initial embeddings of the output embeddings.
This stage allows the updates of both input and output embeddings of the newly added tokens simultaneously.
By aligning between input and output embeddings, the model learns to use the new tokens in both understanding and prediction. 

\noindent\textbf{Stage 4 (all output embeddings):} 
As all the original parameters of the base model were frozen until this stage, we assumed the logits between old and new tokenizers were differently scaled, or less optimized to be used as a whole vocabulary. To this end, we begin to allow the update of the old parameters, specifically the output embeddings of old tokens here, making the model better generate the new tokens. In our preliminary experiments, we found this stage is critical for improving the model's generative capabilities.

\noindent\textbf{Stage 5 (new input and all output embeddings):} At this stage, the training extends to fine-tuning all output embeddings across the model's vocabulary while continuing to refine the input embeddings for the newly added tokens. The goal is to ensure that the model can accurately predict any token within its expanded vocabulary. This phase emphasizes the integration of new tokens within the broader context of the model's linguistic understanding, ensuring that they are both well-represented as inputs and accurately generated as outputs. This dual focus aids in harmonizing the model's overall performance, ensuring that the expanded vocabulary is seamlessly woven into its language generation processes. 

\noindent\textbf{Stage 6 (all layers):} Contrary to being the final phase, this stage represents an advanced step in the vocabulary expansion process, where all model parameters are subject to optimization, including both newly introduced and pre-existing ones. The focus here is on integrating the enhancements made to the embedding layers within the model's overall parameters. Techniques such as QLoRA are utilized not just for efficiency but to ensure the preservation of the model's strong capabilities as much as possible, while allowing effective integration of the expanded vocabulary.

\noindent\textbf{Stage 7 (internal layers):} Following the extensive integration and optimization efforts, this stage serves as a ``cool down'' phase, focusing on updating the model's internal layers, which includes all the layers except the input and output embedding layers. 
The objective is to ensure that the enhancements made during the vocabulary expansion are deeply embedded within the model's core processing capabilities. 
This phase prepares the model for robust performance, ensuring that it not only recognizes and generates the new tokens but does so with a nuanced understanding of their use in varied linguistic contexts.
\section{Implementation Details}
\label{sec:experiment}

\subsection{Datasets}

For \textit{pre-training}, we curated publicly available Korean corpora from diverse sources, such as Korean web content, English vocabulary, and parallel corpus in Korean AI Hub\footnote{\url{https://aihub.or.kr/}}, etc. 
To construct a high-quality pre-training corpus, we applied a set of preprocessing rules: 1) perplexity-based filtering, 2) n-gram repetition~\cite{li2023repetition}-based filtering, and 3) stopword-based filtering. 
For the efficient training of newly added Korean tokens, we intentionally filtered out documents that do not contain many of these tokens. 
Subsequently, we acquired a pre-training corpus totaling 3.2M documents (or, 6.7GB).

\begin{table}[h]
\centering
\resizebox{0.92\columnwidth}{!}{
\begin{tabular}{l|l|l}
\toprule
Model    & \begin{tabular}[c]{@{}c@{}}Total \\ (tokens) \end{tabular} & \begin{tabular}[c]{@{}c@{}}Average \\ (tokens) \end{tabular} \\ \midrule
SOLAR-10.7B    & 3.1B           & 964                        \\ 
EEVE-Korean-10.8B-v1.0  & 1.6B           & 500                        \\ \midrule
Phi-2      & 5.6B           & 1748                       \\ 
EEVE-Korean-2.8B-v1.0    & 1.6B           & 484                        \\ 
\bottomrule
\end{tabular}
}
\caption{Comparison of tokenizers for our 6.7GB pre-training corpus of a total 3.2M documents.}
\label{tab:corpus_tokens}
\end{table}

As can be seen in Table~\ref{tab:corpus_tokens}, for the entire corpus, the SOLAR tokenizer needed to use 3.1B tokens to represent them, but our new tokenizer can do so with almost half, using only 1.6B tokens. This difference becomes even more pronounced in the case of Phi-2 and \texttt{EEVE-Korean-2.8B} models, where they require 5.6B tokens and 1.6B tokens respectively. 
Considering that transformers have a quadratic-increasing computation complexity with respect to token length, this can be interpreted in two significant ways. First, it allows for processing sequences more than 4 times longer on the same GPU. Or second, it means our model can be trained nearly 4 times more computationally efficiently on the same dataset. This difference becomes even more pronounced in the case of the Phi-2 and \texttt{EEVE-Korean-2.8B} tokenizers.

For \textit{fine-tuning} of \texttt{EEVE-Korean} models, we employed the Direct Preference Optimization (DPO;~\citealt{rafailov2023direct}) based on LLaMA-Factory implementation. 
To further enhance the models' capabilities of following Korean instructions, we translated the publicly available instruction datasets, specifically Orca\footnote{\url{https://huggingface.co/datasets/Open-Orca/SlimOrca-Dedup}}~\cite{mukherjee2023orca,SlimOrcaDedup} and UltraFeedback\footnote{\url{https://huggingface.co/datasets/argilla/ultrafeedback-binarized-preferences-cleaned}}~\cite{cui2023ultrafeedback} into Korean. 
In the process of translating these datasets into Korean, ensuring the integrity of programming code formats and correcting translation errors, such as instances where both the source and target languages were inadvertently translated into Korean, was crucial for maintaining the quality and effectiveness of our fine-tuned models. 
We named the fine-tuned models as \texttt{EEVE-Korean-Instruct}. 

\subsection{Training}

As foundational architectures, we opt for SOLAR-10.7B~\cite{kim2023solar} and Phi-2~\cite{li2023textbooks}, because both have shown outstanding performances among similar sizes of LLMs. This choice of foundational architectures aligns with our strategic training objectives, leveraging their proven strengths to ensure our new models achieve similar levels of language understanding and reasoning capabilities in Korean. 

For the training of the model variants, we utilized two distinct codebases: Axolotl\footnote{\url{https://github.com/OpenAccess-AI-Collective/axolotl}} for the initial pre-training phase and LLaMA-Factory\footnote{\url{https://github.com/hiyouga/LLaMA-Factory}}~\cite{llama-factory} for subsequent fine-tuning. These codebases provided a strong and reliable base for our training process.

Specifically, we train our models with a setup of 8 x NVIDIA H100 GPUs with 80GB memory each, utilizing 64 CPU cores. 
For \texttt{EEVE-Korean-10.8B-v1.0}, under bf16 precision, the training process is configured with a sequence of length 4096, gradient accumulation steps set to 4, and a micro-batch size of 8, whereas \texttt{EEVE-Korean-2.8B-v1.0} adopts a sequence length of 2048, gradient accumulation of 16, and a micro-batch size of 16. We employ the AdamW~\cite{loshchilov2018decoupled} optimizer, paired with a cosine learning rate scheduler that includes a warmup phase of 10 steps. The learning rate for the 10.8B variant is set to 4e-5, while we used 2e-4 for the small model. We continued training at each stage until the loss converged, observing the loss converged before reaching 400 global steps, which signifies the efficiency of our training strategy. 
Though our training strategy involves 7 different stages, it is noteworthy that, for our 2.8B variant, the overall pre-training can be done in less than two days as optimizing only the output embeddings doesn't incur much computation. 
\section{Evaluations}
\label{sec:results}

\begin{table*}[t]
  \centering
  % \small % or \footnotesize for smaller size
  \resizebox{0.98\textwidth}{!}{
  \begin{tabular}{@{}lccccccccccc}
  \toprule
  \multirow{3}{*}{Model} & \multirow{3}{*}{Types} & \multicolumn{3}{c}{English} & \multicolumn{5}{c}{Korean} & \multirow{3}{*}{\textbf{Avg.}} \\ \cmidrule(l){3-5} \cmidrule(l){6-10} 
   & & \multirow{2}{*}{\begin{tabular}[c]{@{}c@{}}BQ \\ (0)\end{tabular}} & \multirow{2}{*}{\begin{tabular}[c]{@{}c@{}}CP \\ (0)\end{tabular}} & \multirow{2}{*}{\begin{tabular}[c]{@{}c@{}}HS \\ (0)\end{tabular}} & \multirow{2}{*}{\begin{tabular}[c]{@{}c@{}}BQ \\ (0)\end{tabular}} & \multirow{2}{*}{\begin{tabular}[c]{@{}c@{}}CP \\ (0)\end{tabular}} & \multirow{2}{*}{\begin{tabular}[c]{@{}c@{}}HS \\ (0)\end{tabular}} & \multirow{2}{*}{\begin{tabular}[c]{@{}c@{}}SN \\ (0)\end{tabular}} & \multirow{2}{*}{\begin{tabular}[c]{@{}c@{}}WIC \\ (0)\end{tabular}} & \\ \\ \midrule
  meta-llama/Llama-2-7b-hf & PT & 0.7774 & 0.8700 & 0.5714 & 0.5242 & 0.5700 & 0.4420 & 0.4610 & 0.4881 & 0.5880 \\ 
  meta-llama/Llama-2-13b-hf & PT & 0.8055 & 0.9100 & 0.6006 & 0.5214 & 0.6010 & 0.4380 & 0.5038 & 0.4881 & 0.6086 \\ 
  mistralai/Mistral-7B-v0.1 & PT & 0.8379 & 0.9200 & 0.6129 & 0.6282 & 0.5880 & 0.4300 & 0.5365 & 0.4881 & 0.6302 \\
  meta-llama/Llama-2-7b-chat-hf & FT & 0.7976 & 0.8700 & 0.5779 & 0.5157 & 0.5530 & 0.4160 & 0.4987 & 0.4881 & 0.5896 \\ 
  meta-llama/Llama-2-13b-chat-hf & FT & 0.8165 & 0.8800 & 0.6072 & 0.5057 & 0.5760 & 0.4040 & 0.4685 & 0.4881 & 0.5933 \\
  % maywell/kiqu-70b & PPO & & & & & & & & & \\
  \midrule
  upstage/SOLAR-10.7B-v1.0 (base) & PT & 0.8257 & 0.8700 & 0.6393 & 0.5057 & 0.5750 & 0.4320 & 0.6146 & 0.4881 & 0.6188 \\
  upstage/SOLAR-10.7B-Instruct-v1.0 & FT & \textbf{0.8853} & \textbf{0.9400} & \textbf{0.6866} & 0.8184 & 0.6370 & 0.4560 & 0.5668 & 0.4921 & 0.6853 \\ 
  beomi/OPEN-SOLAR-KO-10.7B$^*$ & PT & 0.8187 & 0.8800 & 0.5570 & 0.8355 & \textbf{0.8010} & \textbf{0.5040} & 0.6952 & 0.4897 & 0.6976 \\ 
  yanolja/EEVE-Korean-10.8B-v1.0$^*$ & PT & 0.8492 & 0.9000 & 0.6203 & 0.8568 & 0.7530 & 0.4900 & 0.6675 & \textbf{0.4992} & 0.7045 \\ 
  yanolja/EEVE-Korean-Instruct-10.8B-v1.0$^*$ & FT & 0.8810 & 0.9300 & 0.6502 & \textbf{0.8860} & 0.7610 & 0.4700 & \textbf{0.9521} & 0.4937 & \textbf{0.7530} \\ 
  \midrule
  microsoft/Phi-2 (base) & PT & \textbf{0.8336} & \textbf{0.9000} & \textbf{0.5583} & 0.5021 & 0.4770 & 0.3280 & 0.5063 & 0.4881 & 0.5742 \\ 
  daekeun-ml/phi-2-ko-v0.1$^*$ & PT & 0.6141 & 0.5800 & 0.3257 & 0.5164 & \textbf{0.6100} & \textbf{0.3860} & 0.4484 & 0.4881 & 0.4961 \\ 
  yanolja/EEVE-Korean-2.8B-v1.0$^*$ & PT & 0.7404 & 0.8900 & 0.5247 & 0.5299 & 0.5820 & 0.3800 & 0.5164 & 0.4881 & 0.5814 \\ 
  yanolja/EEVE-Korean-Instruct-2.8B-v1.0$^*$ & FT & 0.8248 & 0.8700 & 0.5392 & \textbf{0.7066} & 0.5640 & 0.3660 & \textbf{0.5290} & \textbf{0.5230} & \textbf{0.6153} \\  
  \bottomrule
  \end{tabular}
  }
  \caption{Main evaluation results based on \texttt{lm-evaluation-harness}. The dataset names are abbreviated for brevity: BQ for BoolQ, CP for COPA, HS for HellaSwag, and SN for SentiNeg. Korean tasks are from~\cite{jang2022kobest}. Accuracy (\texttt{acc}) is used as the evaluation metric for all tasks. In the `Types' column, the models are categorized into two groups: pre-trained (PT) and fine-tuned (FT). We denote the models trained in Korean datasets with $^*$. For ease of reproduction, we adopt their official names at HuggingFace.}
  \label{tab:main-results}
\end{table*}  

We evaluate our models on both Korean and English LLM benchmarks, to highlight the advantages of our vocabulary expansion method, which could efficiently leverage the strong multilingual capabilities of base foundational models. Desirably, we expect a model to show improved performance in Korean tasks and comparable performance in English tasks. 

\subsection{Benchmarks}

For Korean tasks, we adopt the KoBEST benchmark~\cite{jang2022kobest}, whose tasks are designed to evaluate the various aspects of language understanding and reasoning. Specifically, this benchmark provides a Korean-translated version of language understanding tasks: boolean question answering (BoolQ;~\citealt{clark2019boolq}), commonsense causal reasoning (COPA;~\citealt{roemmele2011choice}, context-sensitive word understanding (WiC;~\citealt{pilehvar2019wic}), commonsense reasoning (HellaSwag;~\citealt{zellers2019hellaswag}), and sentiment negation recognition (SentiNeg). 
For English tasks, we employ the following original tasks of KoBEST, BoolQ, COPA, and HellaSwag, which can better highlight the alignment between the English and Korean capabilities of LLMs. 
To ensure consistent comparisons, we employ an open-source LLM evaluation framework, \texttt{lm-evaluation-harness}\footnote{\url{https://github.com/EleutherAI/lm-evaluation-harness}}~\cite{eval-harness}. 

\subsection{Results}

We now present evaluation results for both our \texttt{EEVE-Korean} and \texttt{EEVE-Korean-Instruct} variants with other top-performing models in Table~\ref{tab:main-results}. 
\texttt{EEVE-Korean-10.8B-v1.0} outperforms other pre-trained models of similar sizes in the average performance. 
It is noteworthy that, \texttt{EEVE-Korean} is the only case where the performance in Korean is improved without compromising the performance in English. 
For example, though OPEN-SOLAR-KO-10.7B, which is built on the same base model as ours, performs slightly better than our \texttt{EEVE-Korean-Instruct-10.8B-v1.0}, it fails to preserve the English capabilities, showing lower performance in English tasks than its base model, SOLAR-10.7B-v1.0.
We observe similar trends even for our smaller model, \texttt{EEVE-Korean-2.8B-v1.0} in comparison with the phi-2-ko-v0.1 model, sharing Phi-2 as its base model.
This demonstrates the effectiveness of our training strategy, especially considering that we used even fewer training tokens than our competitors.

Notably, but not surprisingly, preference tuning on English datasets even makes the models underperform in Korean tasks. For example, LLaMA-2-chat variants, which are the preference-tuned version of LLaMA-2 checkpoints, show improved performances in English tasks (Llama-2-7b 0.7774 $\rightarrow$ Llama-2-7b-chat 0.7976 in English BoolQ), while underperforming in Korean tasks (Llama-2-7b 0.5242 $\rightarrow$ Llama-2-7b-chat 0.5157 in Korean BoolQ), which highlights the importance of Korean-specific training for LLMs. 
On the other hand, we observe that preference tuning our models on Korean instruction datasets doesn't hurt the model performance in English tasks, rather even improving it. We posit that it is because the embedding spaces are already well-aligned between Korean and English tokens, thus fine-tuning on a specific language doesn't incur a significant change in model parameters. 

\section{Conclusion \& Future Work}

The report introduces \texttt{EEVE-Korean-v1.0}, a Korean adaptation of large language models that utilizes an Efficient and Effective Vocabulary Expansion (EEVE) method to enhance Korean text processing capabilities significantly. The method, based on parameter freezing and subword initialization, enables the \texttt{EEVE-Korean-10.8B-v1.0} model to excel in Korean language tasks while maintaining strong English capabilities. Achieved with a corpus of just 2 billion tokens, this approach represents a notable advancement in language model training efficiency and effectiveness. By making these models available to the research community, the project aims to contribute to the development of more inclusive and efficient language processing technologies.

Expanding our vision, future efforts will explore the application of our vocabulary expansion methodology to additional languages, assessing its generalizability and effectiveness. We aim to not only extend the \texttt{EEVE-Korean} model's linguistic range but also to delve deeper into evaluating its reasoning and generative capabilities through diverse tasks, including complex mathematical reasoning tests like GSM8K~\cite{cobbe2021training}, and human evaluations in interactive settings like chatbots~\cite{zheng2023judging}. Moreover, efforts to enhance pre-training data quality, and to analyze performance in code-switching scenarios~\cite{zhang2023multilingual} will underpin our commitment to refining the model's robustness and versatility. These initiatives are designed to broaden the model's applicability and efficacy, pushing the boundaries of what is achievable with advanced language models.

\bibliography{anthology,custom}
\bibstyle{acl_natbib}

\end{document}